\def\year{2018}\relax
\documentclass[letterpaper]{article} %DO NOT CHANGE THIS
\usepackage{aaai18}  %Required
\usepackage{times}  %Required
\usepackage{helvet}  %Required
\usepackage{courier}  %Required
\usepackage{url}  %Required
\usepackage{graphicx}  %Required
\usepackage{amsmath}
\usepackage{subcaption}
\begin{document}
% The file aaai.sty is the style file for AAAI Press 
% proceedings, working notes, and technical reports.
%
\title{Sketch-pix2seq: a Model to Generate Sketches of Multiple Categories}

\author{Yajing Chen$^1$, Shikui Tu$^1$, Yuqi Yi$^1$, Lei Xu$^{1,2,*}$\\
$^1$ Center for Cognitive Machines and Computational Health, \\
and Department of Computer Science and Engineering, Shanghai Jiao Tong University\\
$^2$ Department of Computer Science and Engineering, \\
The Chinese University of Hong Kong\\
\{cyj907,tushikui,awonderfullife,leixu\}@sjtu.edu.cn}

\maketitle
\begin{abstract}
Sketch is an important media for human to communicate ideas, which reflects the superiority of human intelligence. Studies on sketch can be roughly summarized into recognition and generation. Existing models on image recognition failed to obtain satisfying performance on sketch classification. But for sketch generation, a recent study proposed a sequence-to-sequence variational-auto-encoder (VAE) model called sketch-rnn which was able to generate sketches based on human inputs. The model achieved amazing results when asked to learn one category of object, such as an animal or a vehicle. However, the performance dropped when multiple categories were fed into the model. Here, we proposed a model called sketch-pix2seq which could learn and draw multiple categories of sketches. Two modifications were made to improve the sketch-rnn model: one is to replace the bidirectional recurrent neural network (BRNN) encoder with a convolutional neural network(CNN); the other is to remove the Kullback-Leibler divergence from the objective function of VAE. Experimental results showed that models with CNN encoders outperformed those with RNN encoders in generating human-style sketches. Visualization of the latent space illustrated that the removal of KL-divergence made the encoder learn a posterior of latent space that reflected the features of different categories. Moreover, the combination of CNN encoder and removal of KL-divergence, i.e., the sketch-pix2seq model, had better performance in learning and generating sketches of multiple categories and showed promising results in creativity tasks.

\end{abstract}

\section{Introduction}

Back in ancient times, people recorded their life by carving oracle script characters on bones and shells. In modern times, people created art works, e.g., cartoon, to convey information or to express their emotion by simple strokes. Studies related to sketch can be roughly summarized into two categories: recognition and generation. For sketch object recognition, existing computational approaches, including state-of-the-art image recognition models, failed to achieve satisfying performance\cite{netsketch}\cite{drawsketch}. Interestingly, for sketch generation, a model called sketch-rnn\cite{sketchrnn} has made a successful first step.

The sketch-rnn model was proposed to generate sketches based on human-drawn inputs\cite{sketchrnn}. The features learned by the model were represented as a sequence of pen stroke positions. As the raw data were in sequential form, the model used bidirectional recurrent neural network (BRNN) and autoregressive RNN as the encoder and decoder under the framework of Variational AutoEncoder (VAE)\cite{vae}. However, it mainly focused on generating sketches of one category, and the performance for the generation of multiple categories were not satisfactory.

RNN is often used in tasks with time-series data, such as natural language processing\cite{nlp-rnn} and handwriting generation\cite{handwriting}, as it possesses the ability to capture context information and dynamics of time-series data. Together with RNN, convolutional neural network (CNN) is another popular approach in deep learning. A number of CNN models have been proposed to perform image recognition tasks, including the famous AlexNet\cite{AlexNet} and GoogLeNet\cite{GoogleNet}. It is thought that CNN is good at capturing the image local structure.  

Here, we proposed a model called sketch-pix2seq which is capable of learning and generating sketches of multiple categories better than sketch-rnn. Compared with learning one category at a time, learning multiple categories simultaneously has the advantage of saving computational resources. Our work makes two important modifications on the sketch-rnn model, one is to replace the BRNN encoder with a CNN encoder, the other is to remove the enforcement of Gaussian prior on the latent space by ignoring the Kullback-Leibler (KL) divergence in the objective function of VAE. Although utilizing the dynamics and context of strokes by a BRNN encoder reduces the data dimension from a two-dimensional image to one-dimensional sequences, extra information unrelated to the abstract structure of sketches, e.g., the sketching speed, might harm the performance of concept learning. As CNN can capture the local structure of images, which is similar to the ways of human to learn sketch concepts, it might be a better choice for the encoder. Besides, the Gaussian prior enforcement on the latent space of the sketch-rnn model might be unsuitable for multiple categories, because it is unlikely that samples of different categories come from the same distribution. Experimental results show that using CNN as encoders improve the quality of the generated sketches, and the removal of KL-divergence from the cost function allows the encoder to learn a posterior of the latent space that reflects the features of different categories. Latent space interpolation for models without KL-divergence are shown to generate more consistent results than the ones with KL-divergence. Interestingly, the sketch-pix2seq model plays a promising role in creating new sketches by interpolating the latent space of different categories and by feeding cartoon sketches into the model.

\section{Related Work}

Efforts have been made in sketch object recognition. A pioneer study developed a human-drawn sketch database and investigated the sketch recognition abilities of both human and machines\cite{drawsketch}. Another study applied two famous deep learning models used in image recognition, i.e., GoogLeNet\cite{GoogleNet} and AlexNet\cite{AlexNet}, to classify sketches, but their performances were far from satisfactory\cite{netsketch}. However, surprisingly, studies on sketch generation obtained amazing results. A famous research on generating new symbols by one-shot learning showed the potential of machines to achieve human-level learning on sketch-like symbols\cite{Lake1332}. The recently proposed sketch-rnn model was able to produce new sketches based on existing observations\cite{sketchrnn}.

The sketch-rnn is based on Variational AutoEncoder (VAE) framework\cite{vae} which has gained increasing popularity these years. VAE has been applied in generating captions for images\cite{vaecaption}, learning parse trees \cite{grammarvae}, and modeling audience reactions to movies\cite{factorizedvae}. Also, attempts have been made to find a better representation of the latent space for the original VAE. Both the adversarial variational Bayes \cite{adversarial-vb} and adversarial autoencoders \cite{adversarial-vae} applies the idea of generative adversarial network (GAN)\cite{gan} to learn the latent space. In this work, we proposed a VAE-based sketch-pix2seq model which aims to generate sketches of multiple categories.

\section{Methodology}
\subsection{Data set}

The data sets used here came from \emph{QuickDraw}, a public sketch database built by Google. All the sequence data were collected by \emph{The Quick, Draw!}, an online game that requires human to draw a sketch within 20 seconds. The number of training samples for each category is 70,000, while the validation and test samples are both 2,500. We selected six categories to carry out the experiments (Table \ref{tab:data}). The raster images used by the CNN encoder were obtained by first converting the raw sequences to svg files and subsequently converting to monochrome png files of size 48x48.

\subsection{Model}

\begin{figure}[hptb]
    \includegraphics[width=0.47\textwidth]{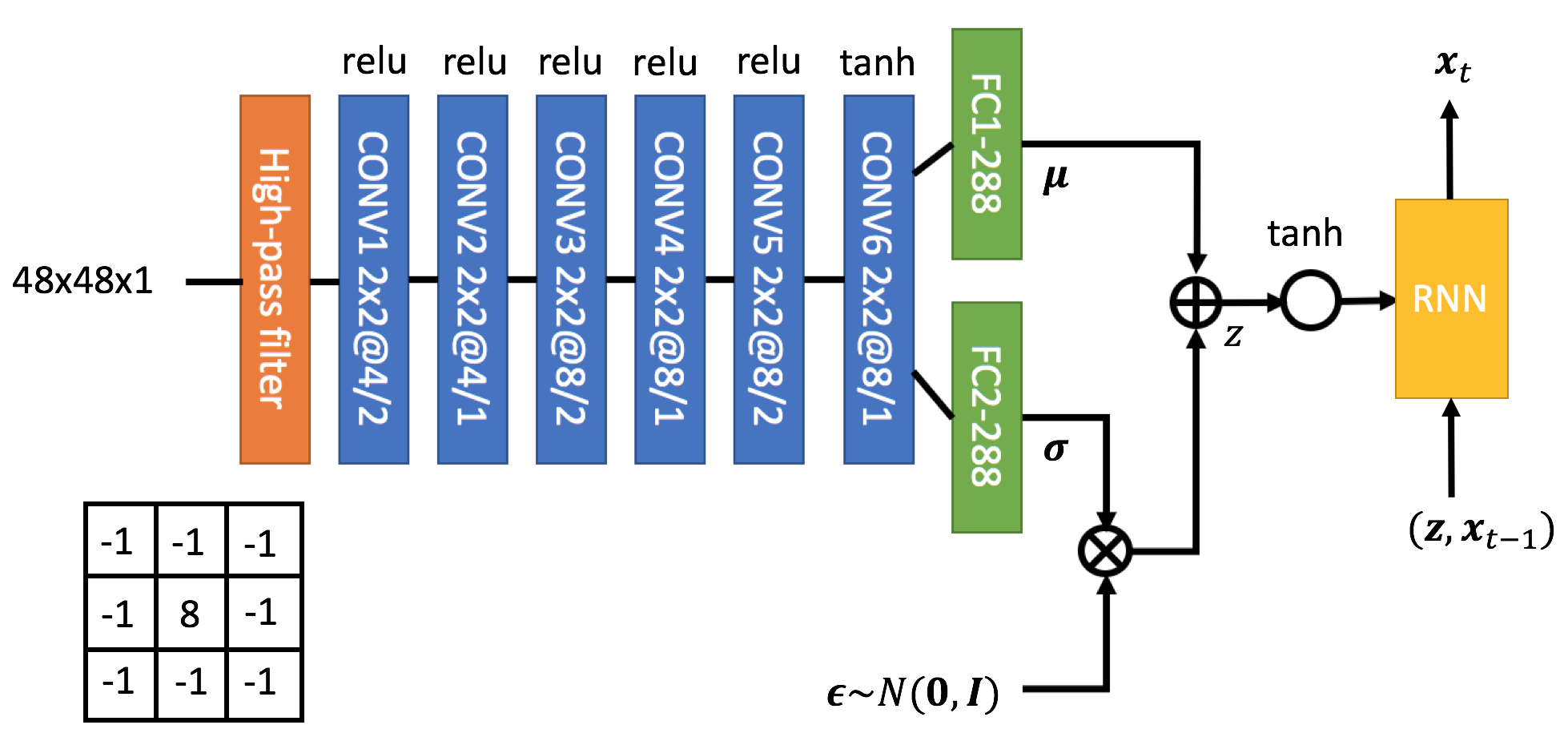}
    \caption{Model Structure. The 3x3 matrix on the bottom left is the high-pass filter applied on the sketch before it is fed into CNN. The convolutional layer configurations are shown as $h\times w$@$d/s$, where $h$, $w$, $d$ and $s$ represent height, width, depth and stride. Above the convolutional layers are the activation functions. The output of the last convolutional layer is rearranged into a one-dimensional vector, which is subsequently fed into two separate fully-connected layers. $\mu$ and $\sigma$ are the mean and standard deviation of the posterior distribution $q_{\phi}(z|X)$ learned by the encoder, where $z=\mu+\sigma \cdot \epsilon$ is the latent vector and $X$ is the input image. $\epsilon$ is a Gaussian noise. $x_t$ is the five-dimensional feature vector at time $t$ in \cite{sketchrnn}. $(z, x_{t-1})$ indicates the concatenation of latent vector and feature vector.}
    \label{fig:model}
\end{figure}

Figure \ref{fig:model} shows the model structure of sketch-pix2seq, which is similar to sketch-rnn, but we made two important modifications. First, the encoder was changed from a bidirectional recurrent neural network (BRNN) to a convolutional neural network (CNN). Second, we no longer enforced a Gaussian prior on the latent space by removing the KL-divergence $D_{KL}(q_{\phi}(z|x)||p_{\theta}(z))$ from the objective function. The reasons for the modifications are as follows.

We adopted CNN as the encoder because it performs well in capturing local structure of images. The learning of sketch concepts for human should be related to the recognition of shapes instead of remembering the sketching process, so CNN seems to be a more suitable option for encoder than BRNN. The decoder of sketch-pix2seq is an autoregressive recurrent neural net as in the sketch-rnn model. 

The original VAE framework seeks to maximize a lower bound $L(\theta,\phi;x)$ of log-likelihood $\log p_{\theta}(x)$:
\begin{multline}
\label{eq:max-log-like}
 L(\theta,\phi;x) = \log p_{\theta}(x) - D_{KL}(q_{\phi}(z|x)||p_{\theta}(z|x)) \\
 \max L(\theta,\phi;x)= \max \{ E_{q_{\phi}(z|x)}[\log p_{\theta}(x|z)] \\ - D_{KL}(q_{\phi}(z|x)||p_{\theta}(z)) \}
\end{multline}

%\begin{equation}
%\label{eq:cost}
%L(\theta,\phi;x) = \log p_{\theta}(x) - D_{KL}(q_{\phi}(z|x)||p_{\theta}(z|x))
%\end{equation}

In the sketch-rnn model, the prior distribution of latent space is assumed as Gaussian $p_{\theta}(z) \sim N(0,I)$, but this assumption might be unsuitable for multiple categories because it is unlikely that data of different categories are drawn from the same distribution. In the VAE framework, latent vectors are actually generated from the posterior $q_{\phi}(z|x)$. Thus, the KL-divergence $D_{KL}(q_{\phi}(z|x)||p_{\theta}(z))$ in the objective function forces all latent vectors to be drawn from the same Gaussian in sketch-rnn, which might be the main factor that brings about unsatisfying results for learning multiple categories. Here, we removed the KL-divergence term from the objective function out of two purposes. One is that we do not bother to specify a reasonable prior for the latent space. The other is that no extra constraint is put on the encoder to learn the posterior $q_{\phi}(z|x)$. As a consequence, the objective function of the sketch-pix2seq model becomes:
\begin{equation}
\label{eq:obj}
    \min E_{q_{\phi}(z|x)}[\log p_{\theta}(x|z)]
\end{equation}

\subsection{Evaluation}
\begin{figure}[hptb]
\centering
\includegraphics[width=0.2\textwidth]{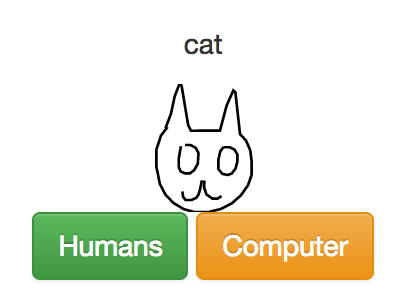}
\caption{Screen shot of the website built for Turing test.}
\label{fig:website}
\end{figure}

To our best knowledge, there exists no standardized criteria to evaluate sketch quality. Thus, we used Turing tests to perform the evaluation tasks. Figure \ref{fig:website} shows the screen shot of the online website built for the tests. During the tests, a model-generated or human-drawn sketch was randomly shown on the screen, and participants were required to tag it as drawn by \emph{Human} or \emph{Computer}.

\section{Experiments}

\subsection{Training details}

We conduct experiments on four models (Table \ref{tab:model}) with two data settings (Table \ref{tab:data}). The parameters used for training models with KL-divergence are the same as sketch-rnn\cite{sketchrnn}. For training models without KL-divergence, we set the KL-weights in sketch-rnn model to be zero, while other parameters are the same.

\begin{table}[htpb]
\centering
\begin{tabular}{l l l}
\hline
Model name & Encoder & Cost function \\
\hline
RNN+KL(sketch-rnn) & BRNN & Equation (\ref{eq:max-log-like}) \\
RNN-KL & BRNN & Equation (\ref{eq:obj}) \\
CNN+KL & CNN & Equation (\ref{eq:max-log-like}) \\
CNN-KL(sketch-pix2seq) & CNN & Equation (\ref{eq:obj}) \\
\hline
\end{tabular}
\caption{Models trained for comparison. RNN+KL: the model with a RNN encoder and KL-divergence; RNN-KL: the model with a RNN encoder and no KL-divergence; CNN+KL: the model with a CNN encoder and KL-divergence; CNN-KL: the model with a CNN encoder and no KL-divergence.}
\label{tab:model}
\end{table}

\begin{table}[htpb]
\centering
\begin{tabular}{c l}
\hline
Setting & Data sets \\
\hline
1 & cat, pig, rabbit \\
2 & cat, pig, rabbit, bus, truck, car \\
\hline
\end{tabular}
\caption{Data Settings}
\label{tab:data}
\end{table}

\subsection{Turing test on the generated sketches}

\begin{figure}[hptb]
    \begin{subfigure}[tpb]{0.47\textwidth}
        \includegraphics[width=\textwidth]{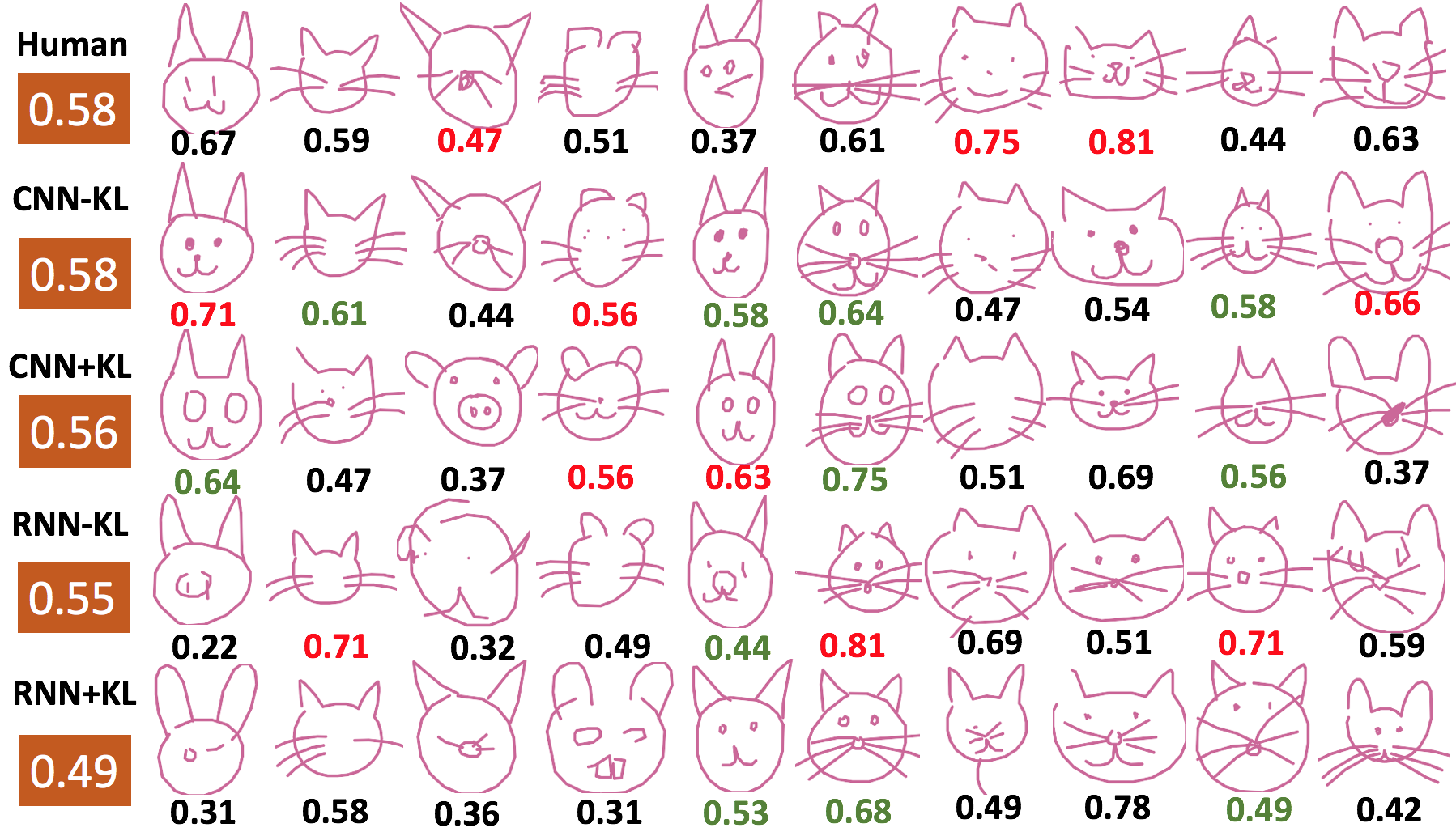}
        \caption{Cat}
        \label{fig:turing_cat}
    \end{subfigure}
    \begin{subfigure}[tpb]{0.47\textwidth}
        \includegraphics[width=\textwidth]{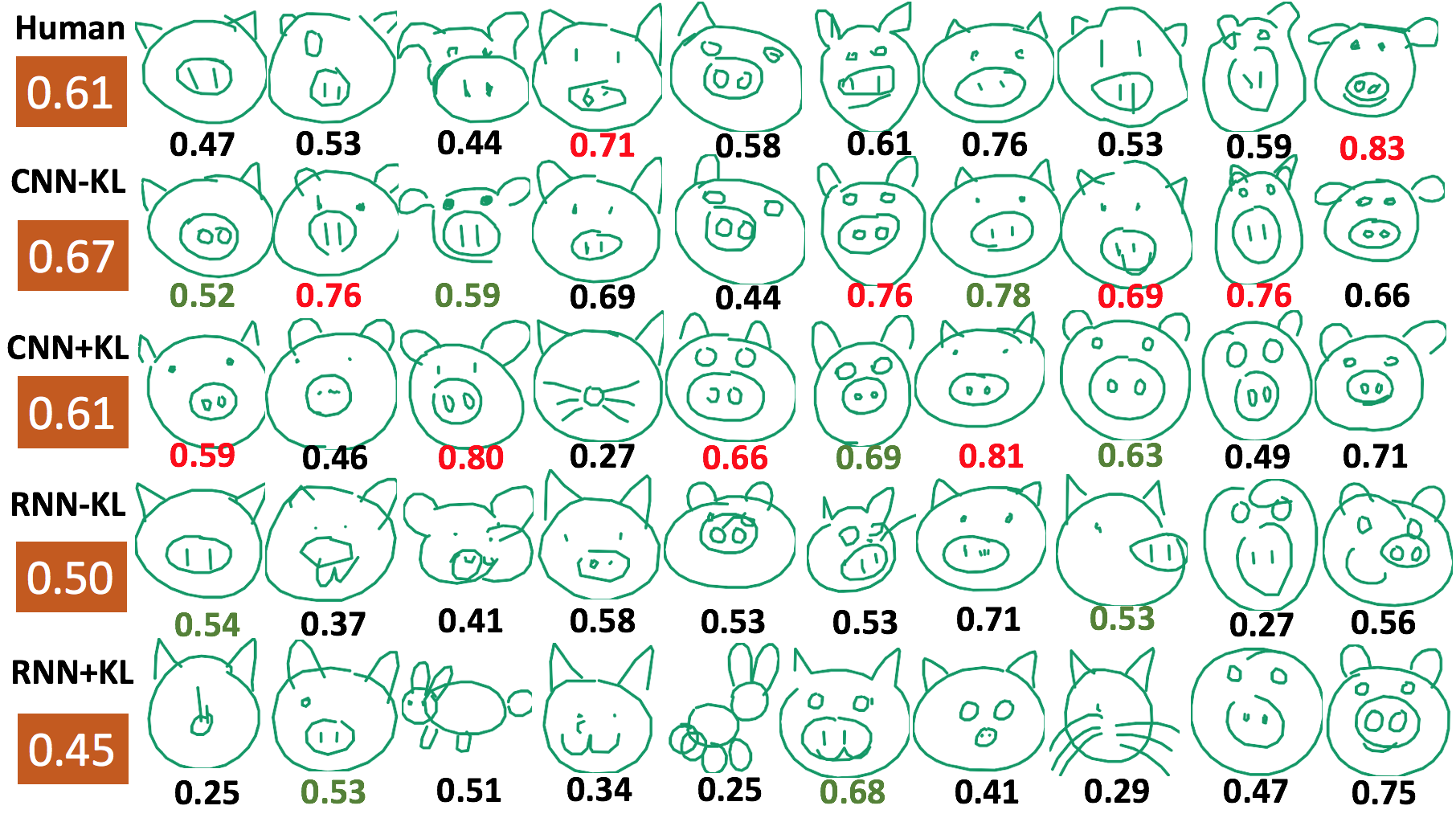}
        \caption{Pig}
        \label{fig:turing_pig}
    \end{subfigure}
    \begin{subfigure}[tpb]{0.47\textwidth}
        \includegraphics[width=\textwidth]{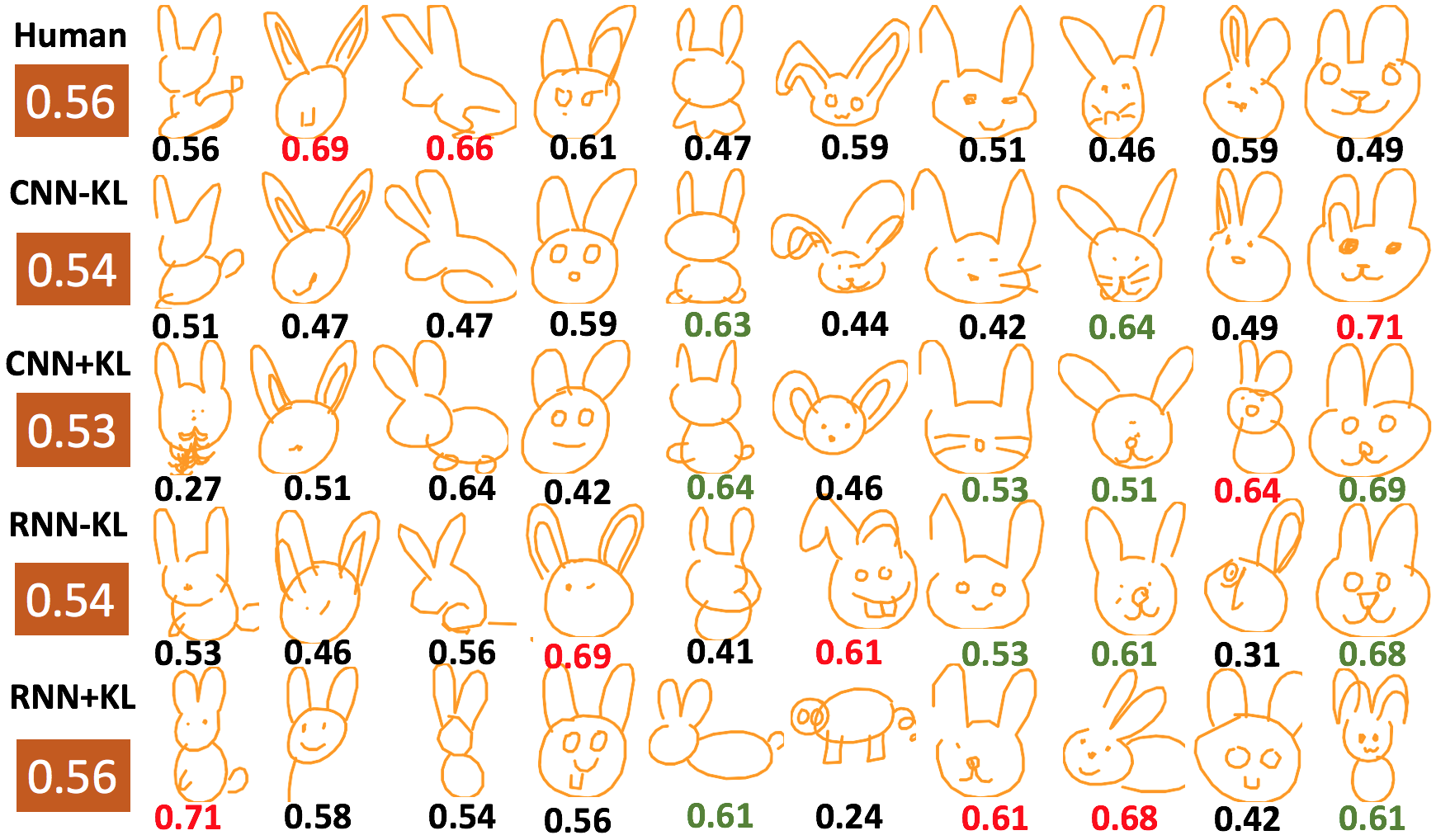}
        \caption{Rabbit}
        \label{fig:turing_rabbit}
    \end{subfigure}
\caption{Sketches used for Turing tests and the resulting statistics. The first row lists the human-drawn inputs, while the following four rows display generated sketches. On the left shows the sketch sources and the average proportion of tagging as human-drawn by human. The number under each sketch is the proportion of people that recognize this sketch as human-drawn.  A green indicates a generated sketch that looks more human-style than the real one, while a red number indicate the most human-style sketch among all sources.}
\label{fig:turing}
\end{figure}

Categories in setting 1 (Table \ref{tab:data}) were used to generate sketches for Turing test. We selected 10 human-drawn sketches for each category. All of them were distinguishable, though some had missing parts. Models in table \ref{tab:model} were trained to generate new sketches based on the 10 selected inputs. In total, 150 sketches (five sources: human, CNN+KL, RNN+KL, CNN-KL, RNN-KL; three categories: cat, pig, rabbit) were used for Turing tests. 61 people participated in the test. After removing 2 people that identified more than 90\% or less than 10\% of the sketches as human-drawn, the results of 59 people were used for further analysis. Figure \ref{fig:turing} lists all the sketches in the test and the resulting statistics. 

The human-drawn sketches in the first row can be counted as a baseline. The sketches from the second to fifth row were the outputs of models based on the human-drawn sketches in the first row. Figure \ref{fig:turing} shows that the performance of different models might be category-related. For cat and rabbit sketches, the difference between four models and human are small. However, when drawing pigs, models with CNN encoders outperform the ones with RNN encoders significantly (0.67 and 0.61 versus 0.50 and 0.45). The overall performance of all models are shown in Table \ref{tab:stats}. 
In general, models with CNN encoders have better performance than those with RNN encoders, and models without KL-divergence also perform better than the ones with KL-divergence. Interestingly, more people identified the sketches generated by the CNN-KL model as human-drawn than the real ones (0.60 versus 0.58). These evidence implies that using CNN as an encoder improves the quality of the generated sketches. 

\begin{table}[hptb]
\centering
\begin{tabular}{l c}
\hline
Model & Human-style \\
\hline
Human & 0.58 \\
CNN-KL & 0.60 \\
CNN+KL & 0.57 \\
RNN-KL & 0.53 \\
RNN+KL & 0.50 \\
\hline
\end{tabular}
\caption{Proportion of sketches identified as human-drawn.}
\label{tab:stats}
\end{table}

Figure \ref{fig:turing} shows that models with KL-divergence failed to distinguish different categories. For example, the RNN+KL model generated rabbits and cats when it was asked to generate pigs, and the CNN+KL model generated a pig when it was asked to generate a cat. These phenomena provide evidence that models with KL-divergence are not suitable for learning multiple categories. In the original work of sketch-rnn\cite{sketchrnn}, the KL-divergence term was recognized as a regularization term. They showed that different weights of the KL-divergence had different influence on the quality of the generated sketches, where smaller weights led to smoother sketches and larger weights produced sketches similar to the inputs. Similar results can be seen in Figure \ref{fig:turing}. Therefore, keeping or removing KL-divergence is a trade-off between quality and accuracy for generated sketches of multiple categories.

\subsection{Visualization of the latent space}

\begin{figure}[hptb]
    \centering
    \includegraphics[width=0.47\textwidth]{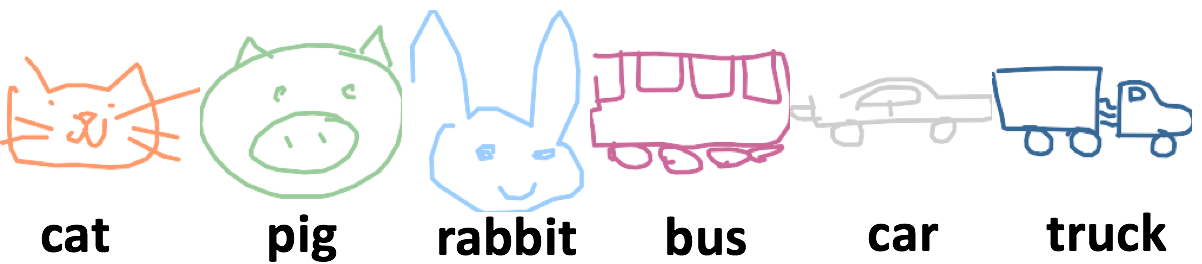}
    \caption{Categories and colors in the latent space graphs.}
    \label{fig:color}
\end{figure}

\begin{figure}[hptb]
    \centering
    \begin{subfigure}[t]{0.47\textwidth}
        \includegraphics[width=\textwidth]{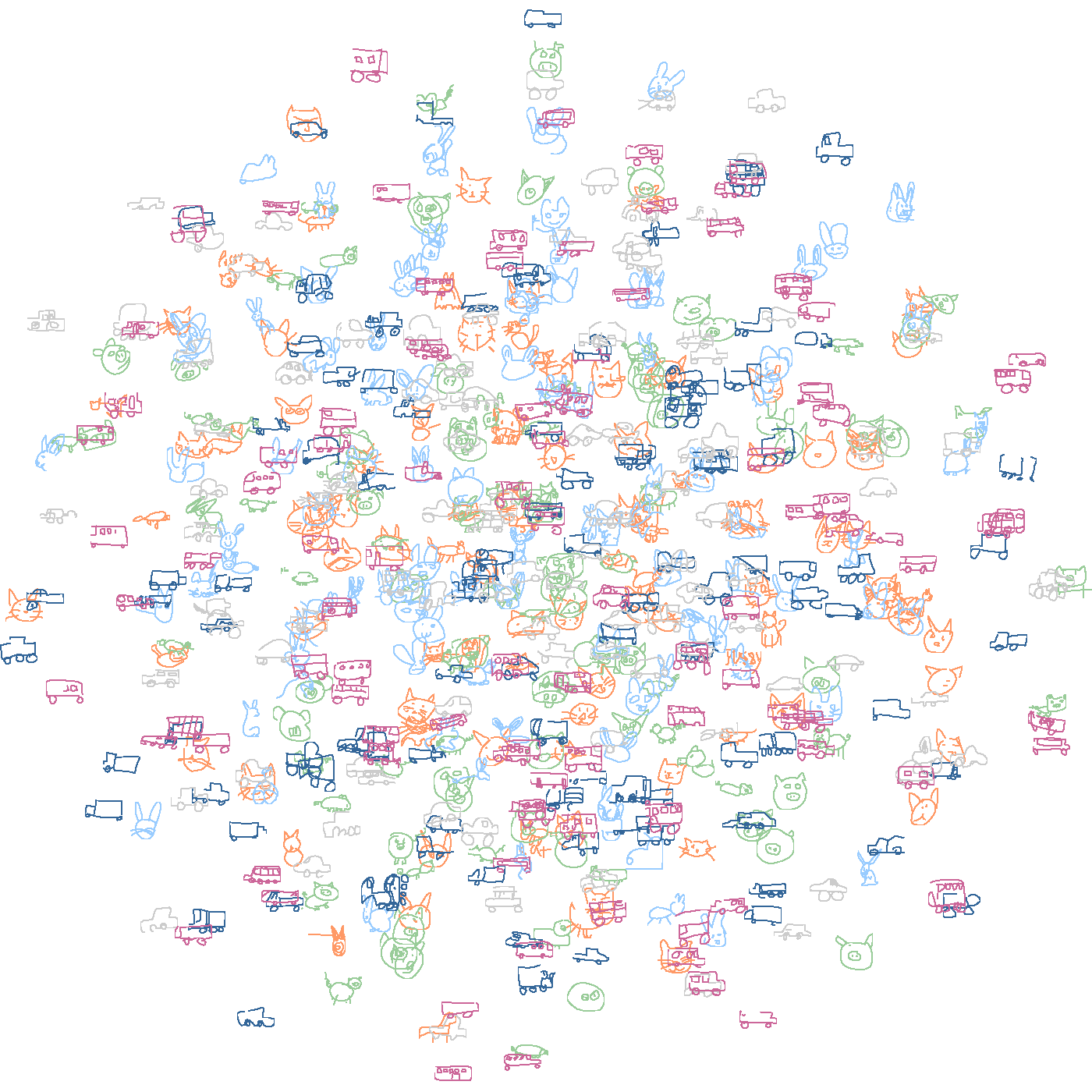}
        \caption{RNN+KL}
        \label{fig:latent_rnn_kl}
    \end{subfigure}
    \begin{subfigure}[t]{0.47\textwidth}
        \includegraphics[width=\textwidth]{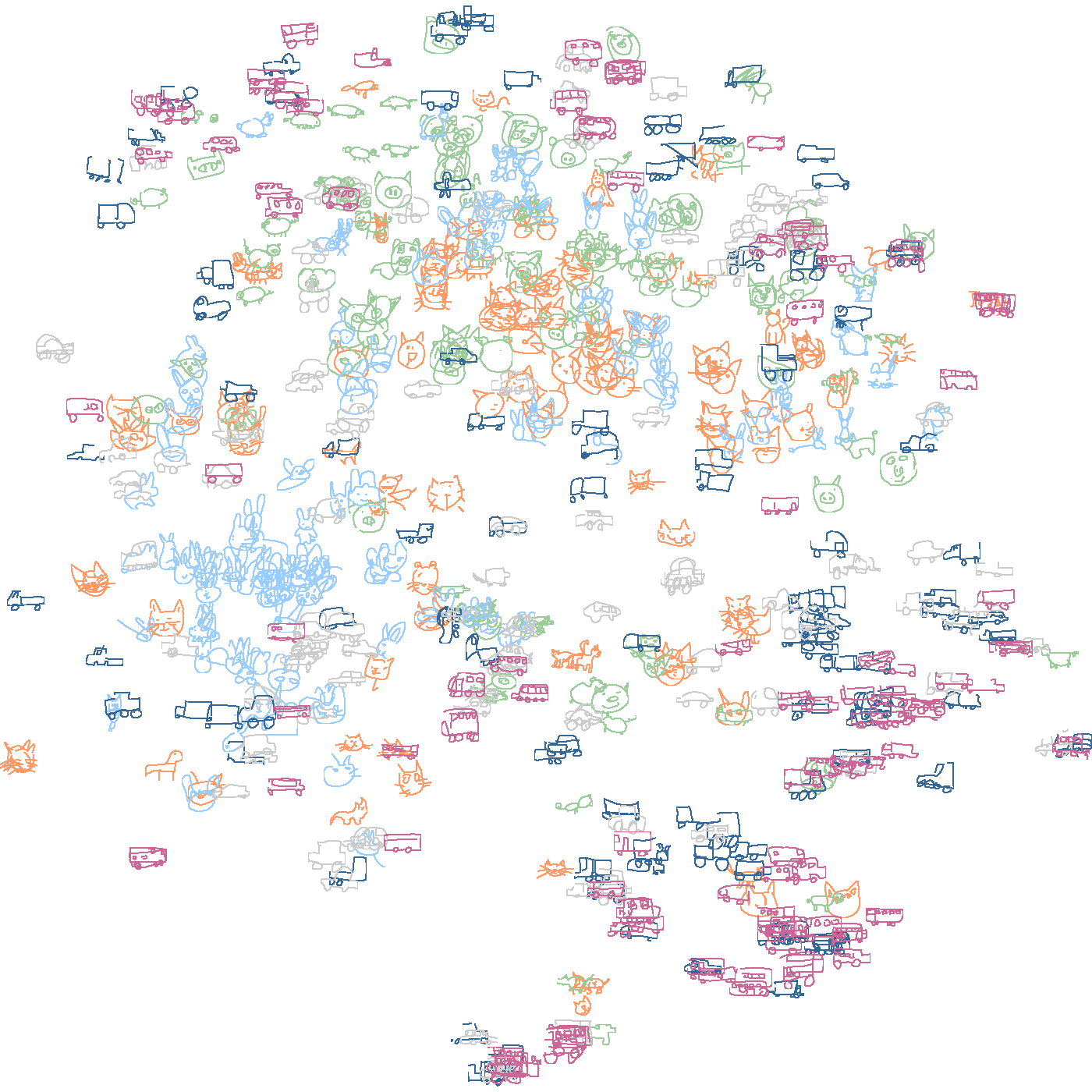}
        \caption{RNN-KL}
        \label{fig:latent_rnn}
    \end{subfigure}
    \caption{Latent spaces for models with RNN as encoders.}
    \label{fig:latent-RNN}
\end{figure}
\begin{figure}[hptb]
    \begin{subfigure}[t]{0.47\textwidth}
        \includegraphics[width=\textwidth]{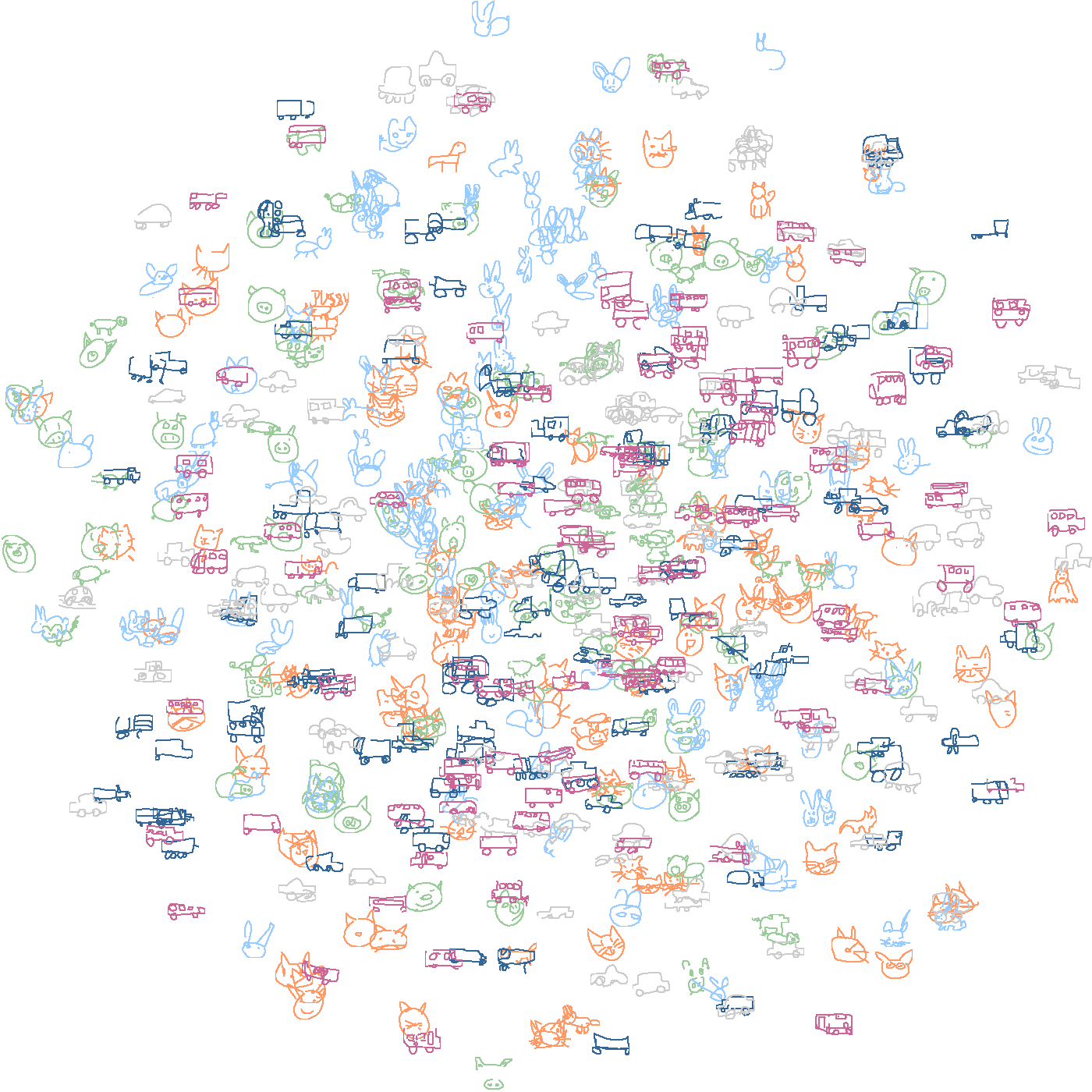}
        \caption{CNN+KL}
        \label{fig:latent_cnn_kl}
    \end{subfigure}
    \begin{subfigure}[t]{0.47\textwidth}
        \includegraphics[width=\textwidth]{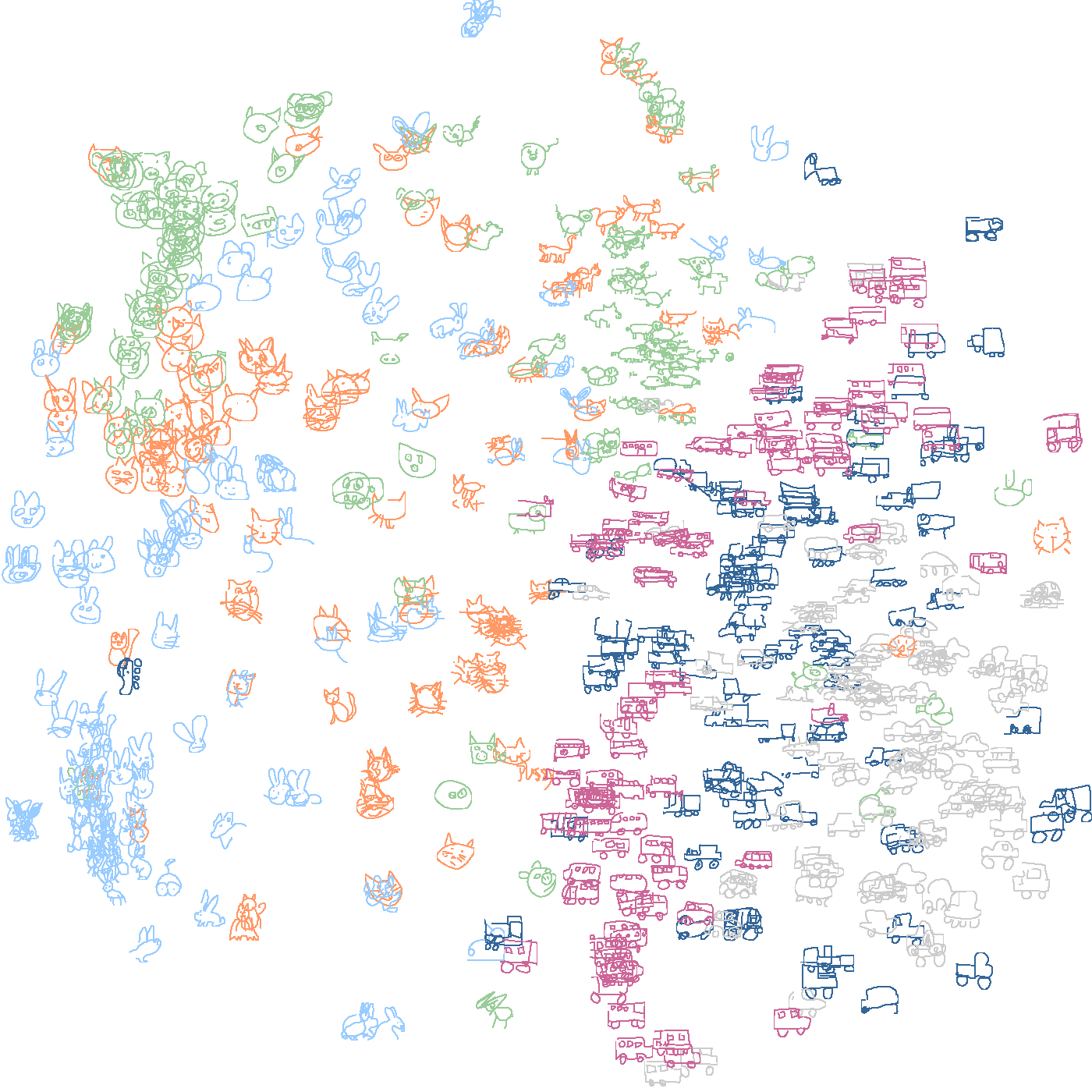}
        \caption{CNN-KL}
        \label{fig:latent_cnn}
    \end{subfigure}
\caption{Latent spaces for models with CNN as encoders.}
\label{fig:latent-CNN}
\end{figure}

\noindent We applied t-SNE clustering \cite{tSNE} to visualize the 128-dimensional latent vectors in a 2D space. Figure \ref{fig:latent-RNN} and \ref{fig:latent-CNN} display the latent spaces of different models trained with setting 2 (see Table \ref{tab:data}). 100 test samples for each category were randomly selected to generate the latent vectors. Figure \ref{fig:color} shows the categories and colors in the latent space graphs. For models with KL-divergence (Figure \ref{fig:latent_rnn_kl} and \ref{fig:latent_cnn_kl}), the latent spaces for different categories are shattered and mixed together. On the contrary, the latent spaces of models without KL-divergence (Figure \ref{fig:latent_rnn} and \ref{fig:latent_cnn}) look non-Gaussian and display clustering effects. In the latent space of RNN-KL, most rabbits are clustered on the bottom left, while most pigs and cats are clustered on the top right. Most of the vehicles are gathered on the bottom right. In the latent space of CNN-KL, the clustering effects for different categories are more distinguishable than RNN-KL, where the animals are clustered on the left, and the vehicles are clustered on the right. Moreover, a clear separation can be seen between different vehicles. The latent space configuration of CNN-KL indicates that CNN encodes the input sketches according to their shapes, since samples that look similar are clustered together.

\subsection{Latent Space Interpolation}

We performed linear interpolation on the latent vectors of two categories for different models. All models listed in Table \ref{tab:model} were trained with setting 2 of Table \ref{tab:data}. The sketches used for interpolation were shown in Figure \ref{fig:color}.

\begin{figure}[hptb]
    \begin{subfigure}[t]{0.47\textwidth}
        \includegraphics[width=\textwidth]{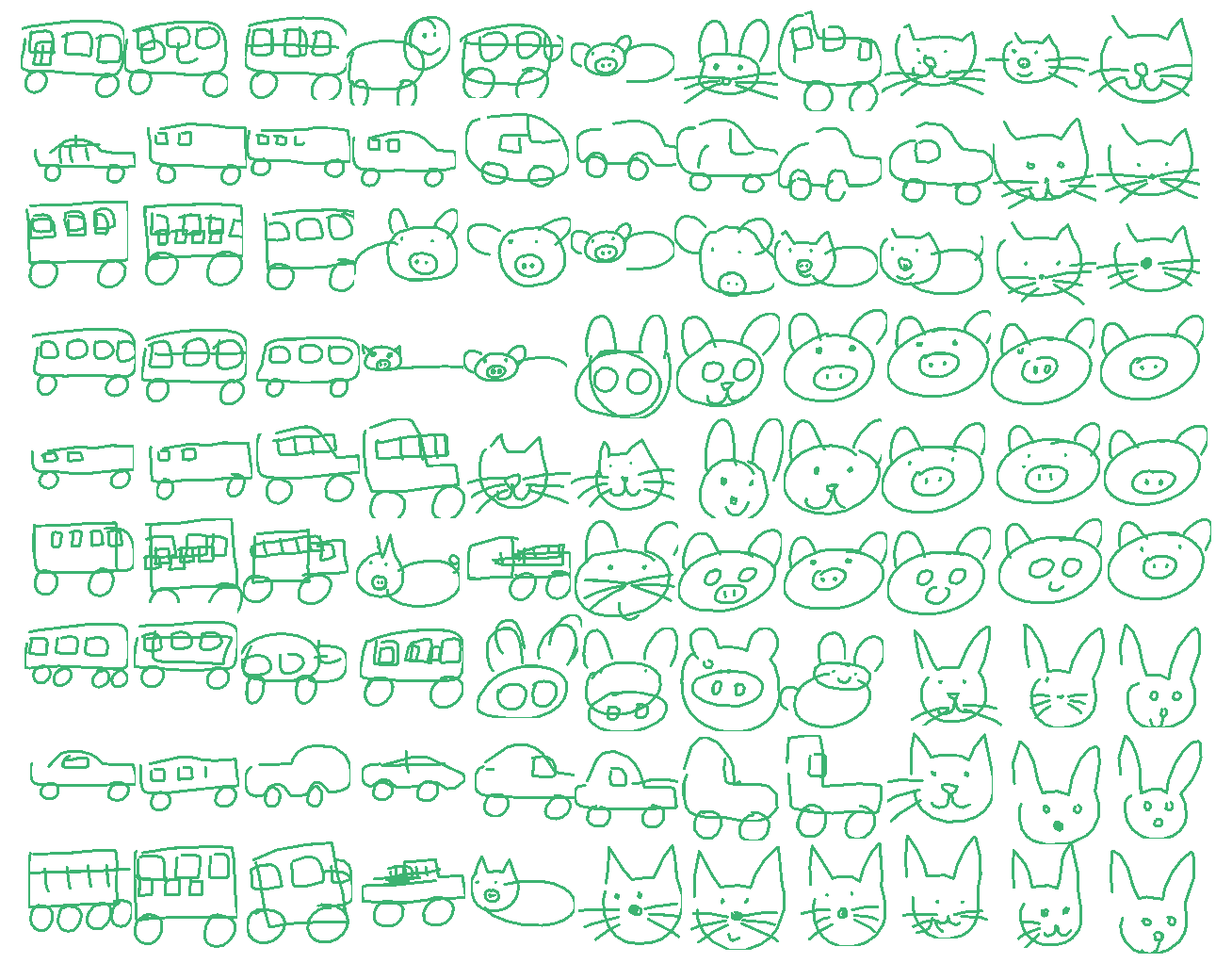}
        \caption{RNN+KL}
        \label{fig:intpl-RNN+KL}
    \end{subfigure}
    \begin{subfigure}[t]{0.47\textwidth}
        \includegraphics[width=\textwidth]{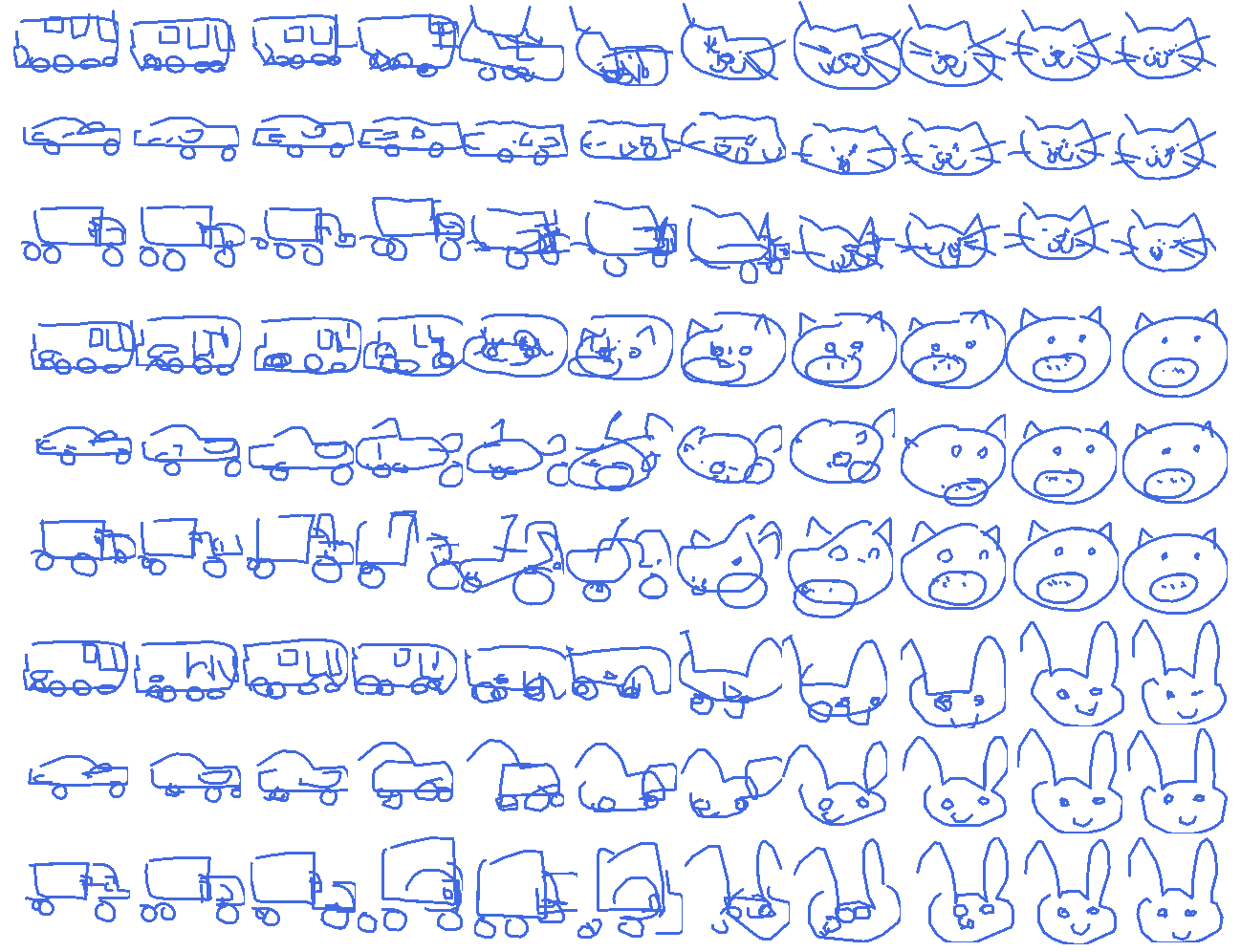}
        \caption{RNN-KL}
        \label{fig:intpl-RNN-KL}
    \end{subfigure}
    \caption{Interpolation for models with RNN encoders.}
    \label{fig:intpl-RNN}
\end{figure}

\begin{figure}[hptb]
    \begin{subfigure}[t]{0.47\textwidth}
        \includegraphics[width=\textwidth]{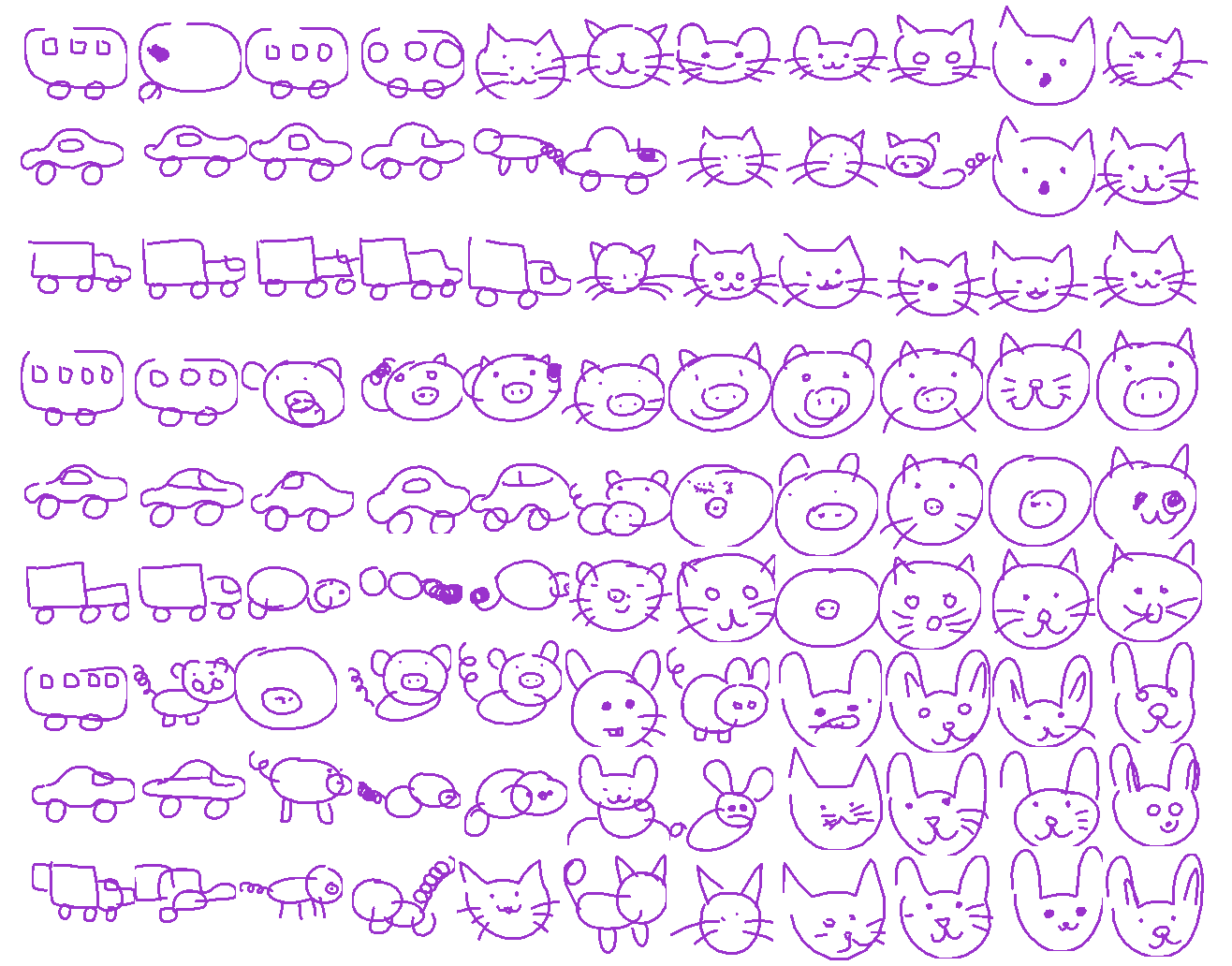}
        \caption{CNN+KL}
        \label{fig:intpl-CNN+KL}
    \end{subfigure}
    \begin{subfigure}[t]{0.47\textwidth}
        \includegraphics[width=\textwidth]{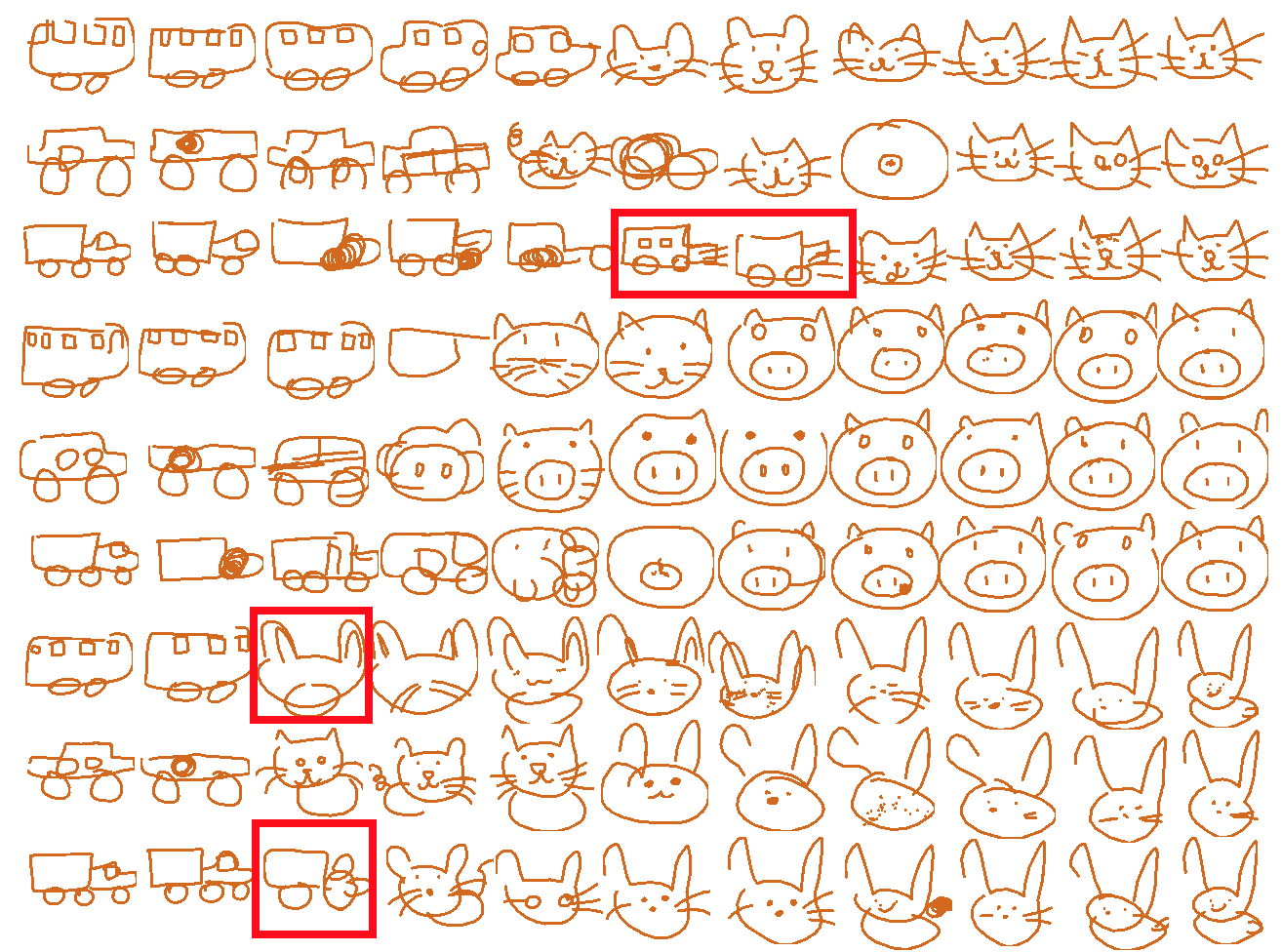}
        \caption{CNN-KL. The red rectangles indicate the interesting and meaningful results found by interpolation.}
        \label{fig:intpl-CNN-KL}
    \end{subfigure}
    \caption{Interpolation for models with CNN encoders.}
    \label{fig:intpl-CNN}
\end{figure}

\begin{figure}[hptb]
    \centering
    \includegraphics[width=0.3\textwidth]{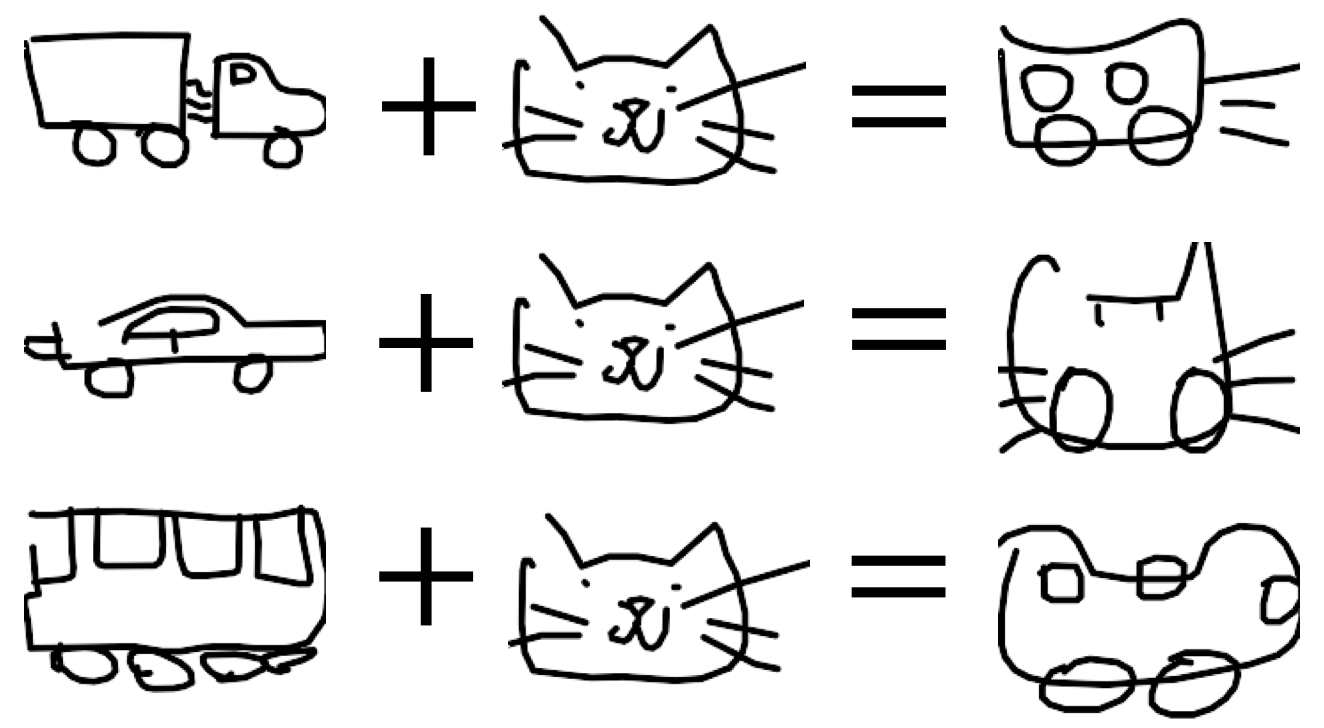}
    \caption{Interpolation of vehicles and cat by sketch-pix2seq.}
    \label{fig:intpl-vehicle-cat}
\end{figure}

Figure \ref{fig:intpl-RNN} and Figure \ref{fig:intpl-CNN} display the interpolation results by different models. We applied linear interpolation of $z = w_1 z_1 + w_2 z_2$ on latent vectors, where $w_1$ and $w_2$ are interpolation weights, $z_1$ and $z_2$ are latent vectors of different categories, and $z$ is the output latent vector. The interpolation weights should satisfy the constraint $w_1 + w_2 = 1$. Both $w_1$ and $w_2$ range from 0 to 1 by step 0.1, resulting in 11 generated sketches in each row. The categories used for interpolation in row order are: bus-cat, car-cat, truck-cat, bus-pig, car-pig, truck-pig, bus-rabbit, car-rabbit and truck-rabbit. 

The difference of the four models can be inferred from the interpolation results. The RNN+KL model not only had trouble distinguishing different categories, but also generated unexpected categories in the middle of interpolation. In Col 1 of Figure \ref{fig:intpl-RNN+KL}, the model generated buses whenever it was asked to draw a truck (in Row 3,6,9). In Row 1, a rabbit and a pig appear where the two categories used for interpolation are buses and cats. The CNN+KL model (Figure \ref{fig:intpl-RNN+KL}) also produced unexpected categories during interpolation. In Row 7 where a bus and a rabbit are interpolated, pigs show up for multiple times. Compared with models with KL-divergence, models without KL-divergence perform much better during interpolation. The transformation between categories looks smooth and consistent in both RNN-KL and CNN-KL (Figure \ref{fig:intpl-RNN-KL} and \ref{fig:intpl-CNN-KL}). The differences between neighboring sketches of RNN-KL model show that the transformation of sketches during interpolation is about the style, orientation and configuration of strokes (Figure \ref{fig:intpl-RNN-KL}). On the other hand, the differences between neighboring sketches of CNN-KL model show that the transformation is about the shape of sketches.

In sketch-rnn, experiments were also conduct on interpolation. They trained models using two data sets and compared the interpolation results for different weights of KL-divergence. They showed that the generated sketches were more coherent for larger weights of KL-divergence. However, the coherency for models with KL-divergence no long exists when they are trained with more than three data sets according to our results, because sketches of an unexpected third category are likely to show up. These evidence might imply that models with KL-divergence should be used for training two data sets at most, while models without KL-divergence are suitable for multiple data sets.

Furthermore, the interpolation results of the CNN-KL model show its potential to generate creative sketches by linear interpolation (see the red rectangles in Figure \ref{fig:intpl-CNN-KL}). In Row 3, the truck-cat interpolation produces a truck with whiskers at the front. In Row 7, the bus-rabbit interpolation generates a vehicle with a single wheel and a shield similar to a rabbit ear at either side. In Row 9, the truck-rabbit interpolation brings about a truck with a rabbit head. To further investigate the potential of the CNN-KL model on creativity, we interpolated different vehicles with a cat with smaller granularity of weight ranges (from 0 to 1 at step 0.02). Figure \ref{fig:intpl-vehicle-cat} shows that the truck-cat interpolation generates a vehicle with a face-like body and whiskers at the front; the car-cat interpolation outputs a cat with two wheels on its face; the bus-cat interpolation produces a bus with a cat-face body and a light on the top. All these generated sketches were never seen in the training samples, but the sketch-pix2seq model successfully combined the shape features of different categories and created a new sketch.

\subsection{Cartoon figures as inputs}
\begin{figure}[hptb]
\centering
\includegraphics[width=0.47\textwidth]{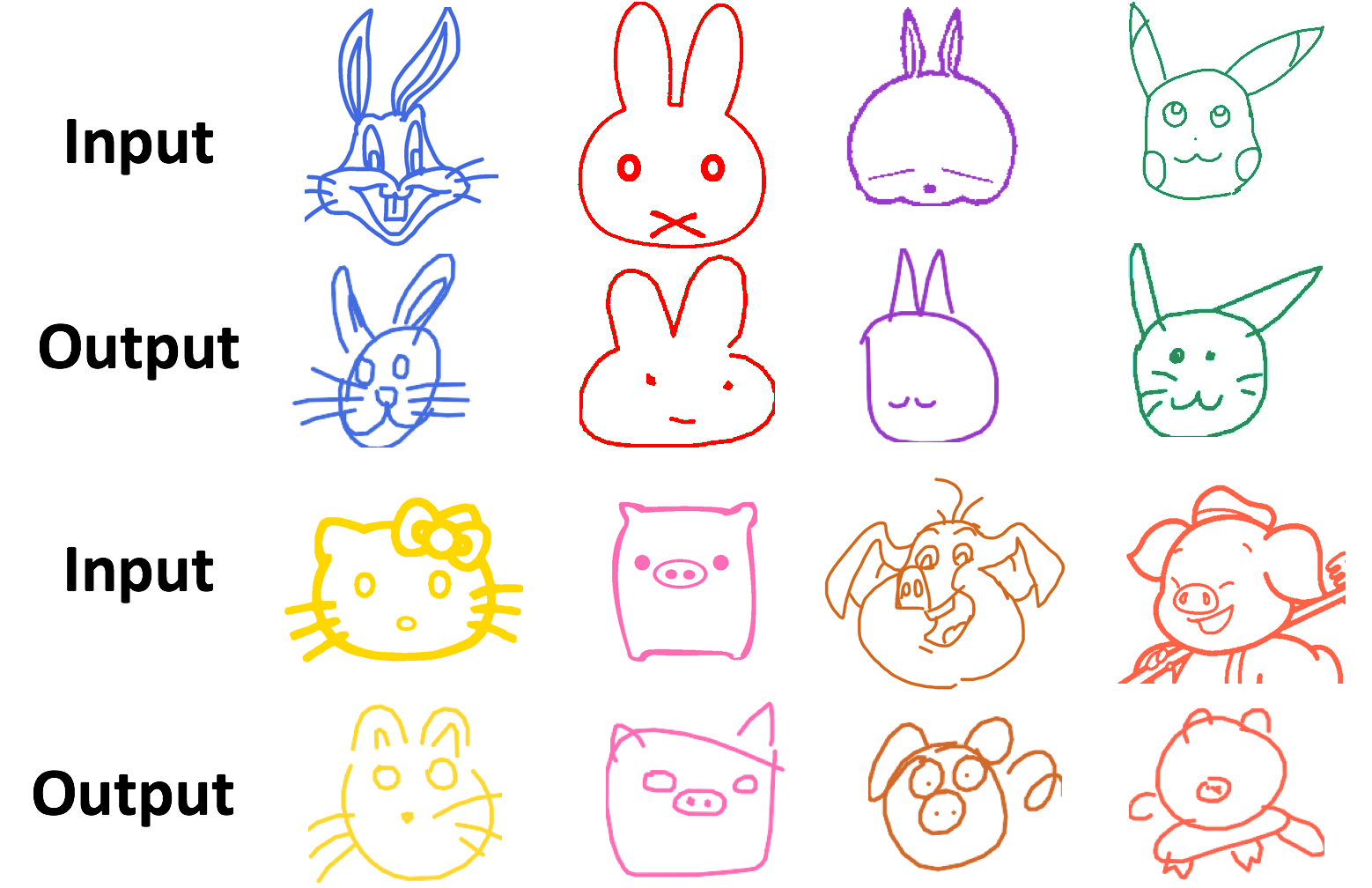}
\caption{Cartoon figure inputs and the generated sketches. The cartoon figures in Row 1: Bugs Bunny, Miffy, Mashimaro and Pikachu. The cartoon figures in Row 3: Hello Kitty, Monokuro Boo, Zhu Bajie and Zhu Bajie.}
\label{fig:cartoon}
\end{figure}

To further investigate the generalization ability of the sketch-pix2seq model, we fed several famous cartoon sketches into the model and evaluated the outputs. Figure \ref{fig:cartoon} shows the input cartoon figures and the corresponding generated sketches. The cartoon prototypes include pigs, rabbits and cats: Zhu Bajie originates from pigs; Bugs Bunny, Miffy and Mashimaro originate from rabbits; Hello Kitty originates from cats. As Pikachu has two long ears that look like the ones of rabbits, it was also selected for experiment. The generated sketches in Figure \ref{fig:cartoon} show the strong capability of sketch-pix2seq model in recognizing categories. Though there exists significant difference between the cartoon figure inputs and the generated sketches, their styles are similar. For example, the generated rabbit from Pikachu not only has a pair of long ears whose orientations are exactly the same as Pikachu, but also has a similar smiley facial expression.

\section{Conclusion}

In this paper, we focus on the generation of sketches for multiple categories. Though sketch-rnn\cite{sketchrnn} was able to produce amazing results when learning one category, the quality of the generated sketches dropped with the increasing number of categories. To resolve this problem, we proposed a VAE-based sketch-pix2seq model. Two important modifications are made here: one is that the encoder is changed to a CNN; the other is that the KL-divergence $D_{KL}(q_{\phi}(z|x)||p_{\theta}(z|x))$ is removed from the cost function.

We compared the performance of four models, i.e., RNN encoder with KL-divergence (RNN+KL), RNN encoder without KL-divergence (RNN-KL), CNN encoder with KL-divergence (CNN+KL) and CNN encoder without KL-divergence (CNN-KL). Turing tests were carried out to evaluate the quality of the generated sketches. The results showed that models with CNN encoders performed better in generating human-style sketches than RNN encoders. Besides, models without KL-divergence generated sketches with higher accuracy than those with KL-divergence. 

Visualization of the latent space shows the scattered and mixed structure for models with KL-divergence, while the latent space distributions for models without KL-divergence are clustered according to the features of different categories. The latent space configuration explained why models with KL-divergence tend to generate sketches of wrong categories compared with models without KL-divergence.

The latent space interpolation of different categories indicates that models with KL-divergence tend to produce sketches of an unexpected third category during interpolation, which again implies that models with KL-divergence are unsuitable for learning multiple categories. Moreover, the interpolation results of RNN-KL and CNN-KL show that RNN encoders learn stroke-related features, while CNN encoders learn shape-related features. Besides, we can see the potential of sketch-pix2seq in generating creative sketches by simple linear interpolation. Interestingly, the model can also generate sketches based on cartoon figures and similar styles are shared between inputs and outputs. 

In conclusion, we think that the sketch-pix2seq model is not only suitable for generating sketches of multiple categories, but also promising for creativity tasks.
\subsection{Acknowledgements.}
The author would like to thank Zhang Tianqi, Li Yang, Huang Hongru and Zhou Xuanyi for their help.

%\nocite{langley00}

\bibliography{example_paper}

\begin{thebibliography}{}

\bibitem[\protect\citeauthoryear{Ballester and de Ara{\'u}jo}{2016}]{netsketch}
Ballester, P., and de~Ara{\'u}jo, R.~M.
\newblock 2016.
\newblock On the performance of googlenet and alexnet applied to sketches.
\newblock In {\em AAAI},  1124--1128.

\bibitem[\protect\citeauthoryear{Deng \bgroup et al\mbox.\egroup
  }{2017}]{factorizedvae}
Deng, Z.; Navarathna, R.; Carr, P.; Mandt, S.; Yue, Y.; Matthews, I.; and Mori,
  G.
\newblock 2017.
\newblock Factorized variational autoencoders for modeling audience reactions
  to movies.

\bibitem[\protect\citeauthoryear{Eitz, Hays, and Alexa}{2012}]{drawsketch}
Eitz, M.; Hays, J.; and Alexa, M.
\newblock 2012.
\newblock How do humans sketch objects?
\newblock {\em ACM Trans. Graph.} 31(4):44--1.

\bibitem[\protect\citeauthoryear{Goodfellow \bgroup et al\mbox.\egroup
  }{2014}]{gan}
Goodfellow, I.; Pouget-Abadie, J.; Mirza, M.; Xu, B.; Warde-Farley, D.; Ozair,
  S.; Courville, A.; and Bengio, Y.
\newblock 2014.
\newblock Generative adversarial nets.
\newblock In {\em Advances in neural information processing systems},
  2672--2680.

\bibitem[\protect\citeauthoryear{{Graves}}{2013}]{handwriting}
{Graves}, A.
\newblock 2013.
\newblock {Generating Sequences With Recurrent Neural Networks}.
\newblock {\em ArXiv e-prints}.

\bibitem[\protect\citeauthoryear{{Ha} and {Eck}}{2017}]{sketchrnn}
{Ha}, D., and {Eck}, D.
\newblock 2017.
\newblock {A Neural Representation of Sketch Drawings}.
\newblock {\em ArXiv e-prints}.

\bibitem[\protect\citeauthoryear{{Kingma} and {Welling}}{2013}]{vae}
{Kingma}, D.~P., and {Welling}, M.
\newblock 2013.
\newblock {Auto-Encoding Variational Bayes}.
\newblock {\em ArXiv e-prints}.

\bibitem[\protect\citeauthoryear{Krizhevsky, Sutskever, and
  Hinton}{2012}]{AlexNet}
Krizhevsky, A.; Sutskever, I.; and Hinton, G.~E.
\newblock 2012.
\newblock Imagenet classification with deep convolutional neural networks.
\newblock In {\em Advances in neural information processing systems},
  1097--1105.

\bibitem[\protect\citeauthoryear{Kusner, Paige, and
  Hern{\'a}ndez-Lobato}{2017}]{grammarvae}
Kusner, M.~J.; Paige, B.; and Hern{\'a}ndez-Lobato, J.~M.
\newblock 2017.
\newblock Grammar variational autoencoder.
\newblock {\em arXiv preprint arXiv:1703.01925}.

\bibitem[\protect\citeauthoryear{Lake, Salakhutdinov, and
  Tenenbaum}{2015}]{Lake1332}
Lake, B.~M.; Salakhutdinov, R.; and Tenenbaum, J.~B.
\newblock 2015.
\newblock Human-level concept learning through probabilistic program induction.
\newblock {\em Science} 350(6266):1332--1338.

\bibitem[\protect\citeauthoryear{Maaten and Hinton}{2008}]{tSNE}
Maaten, L. v.~d., and Hinton, G.
\newblock 2008.
\newblock Visualizing data using t-sne.
\newblock {\em Journal of Machine Learning Research} 9(Nov):2579--2605.

\bibitem[\protect\citeauthoryear{Makhzani \bgroup et al\mbox.\egroup
  }{2015}]{adversarial-vae}
Makhzani, A.; Shlens, J.; Jaitly, N.; Goodfellow, I.; and Frey, B.
\newblock 2015.
\newblock Adversarial autoencoders.
\newblock {\em arXiv preprint arXiv:1511.05644}.

\bibitem[\protect\citeauthoryear{Mescheder, Nowozin, and
  Geiger}{2017}]{adversarial-vb}
Mescheder, L.; Nowozin, S.; and Geiger, A.
\newblock 2017.
\newblock Adversarial variational bayes: Unifying variational autoencoders and
  generative adversarial networks.
\newblock {\em arXiv preprint arXiv:1701.04722}.

\bibitem[\protect\citeauthoryear{Pu \bgroup et al\mbox.\egroup
  }{2016}]{vaecaption}
Pu, Y.; Gan, Z.; Henao, R.; Yuan, X.; Li, C.; Stevens, A.; and Carin, L.
\newblock 2016.
\newblock Variational autoencoder for deep learning of images, labels and
  captions.
\newblock In {\em Advances in Neural Information Processing Systems},
  2352--2360.

\bibitem[\protect\citeauthoryear{Socher \bgroup et al\mbox.\egroup
  }{2011}]{nlp-rnn}
Socher, R.; Lin, C.~C.; Manning, C.; and Ng, A.~Y.
\newblock 2011.
\newblock Parsing natural scenes and natural language with recursive neural
  networks.
\newblock In {\em Proceedings of the 28th international conference on machine
  learning (ICML-11)},  129--136.

\bibitem[\protect\citeauthoryear{{Szegedy} \bgroup et al\mbox.\egroup
  }{2014}]{GoogleNet}
{Szegedy}, C.; {Liu}, W.; {Jia}, Y.; {Sermanet}, P.; {Reed}, S.; {Anguelov},
  D.; {Erhan}, D.; {Vanhoucke}, V.; and {Rabinovich}, A.
\newblock 2014.
\newblock {Going Deeper with Convolutions}.
\newblock {\em ArXiv e-prints}.

\end{thebibliography}
\bibliographystyle{aaai}

\end{document}